\documentclass[letterpaper]{article} 
\usepackage{aaai2026}  
\usepackage{times}  
\usepackage{helvet}  
\usepackage{courier}  
\usepackage[hyphens]{url}  
\usepackage{graphicx} 
\usepackage{natbib}  
\usepackage{caption} 
\usepackage{algorithm}
\usepackage{algorithmic}

\usepackage{multirow}
\usepackage{makecell}
\usepackage{listings}
\usepackage{longtable}
\usepackage{booktabs}
\usepackage{newfloat}
\usepackage{listings}
\DeclareCaptionStyle{ruled}{labelfont=normalfont,labelsep=colon,strut=off} 
\floatstyle{ruled}
\newfloat{listing}{tb}{lst}{}
\floatname{listing}{Listing}
\usepackage{xurl}
\title{\textsc{LocalBench}: Benchmarking LLMs on County-Level Local Knowledge and Reasoning}
\author{Zihan Gao\textsuperscript{\rm 1}, Yifei Xu\textsuperscript{\rm 2}, Jacob Thebault-Spieker\textsuperscript{\rm 1}}
\affiliations{
    \textsuperscript{\rm 1}University of Wisconsin-Madison\\
    \textsuperscript{\rm 2}University of California, Los Angeles \\
    zihan.gao@wisc.edu, yxu@cs.ucla.edu, jacob.thebaultspieker@wisc.edu
}

\usepackage{bibentry}

\begin{document}

\maketitle

\begin{abstract}
Large language models (LLMs) have been widely evaluated on macro-scale geographic tasks, such as global factual recall, event summarization, and regional reasoning. Yet, their ability to handle hyper-local knowledge remains poorly understood. This gap is increasingly consequential as real-world applications, from civic platforms to community journalism, demand AI systems that can reason about neighborhood-specific dynamics, cultural narratives, and local governance. Existing benchmarks fall short in capturing this complexity, often relying on coarse-grained data or isolated references.
We present \textsc{LocalBench}, the first benchmark designed to systematically evaluate LLMs on county-level local knowledge across the United States. Grounded in the Localness Conceptual Framework, \textsc{LocalBench} includes 14,782 validated question-answer pairs across 526 U.S. counties in 49 states, integrating diverse sources such as Census statistics, local subreddit discourse, and regional news. It spans physical, cognitive, and relational dimensions of locality.
Using \textsc{LocalBench}, we evaluate 13 state-of-the-art LLMs under both closed-book and web-augmented settings. Our findings reveal critical limitations: even the best-performing models reach only $56.8\%$ accuracy on narrative-style questions and perform below $15.5\%$ on numerical reasoning. Moreover, larger model size and web augmentation do not guarantee better performance, for example, search improves Gemini's accuracy by $+13.6\%$, but reduces GPT-series performance by $-11.4\%$. These results underscore the urgent need for language models that can support equitable, place-aware AI systems: capable of engaging with the diverse, fine-grained realities of local communities across geographic and cultural contexts.
\end{abstract}

\begin{links}
    \link{Datasets}{https://github.com/MadCollab/LocalBench}
\end{links}

\section{Introduction}

LLMs have been extensively evaluated on tasks involving macro-scale geographic knowledge, such as factual recall~\cite{moayeri2024worldbench}, global event summarization~\cite{almeida2025tiebe}, and cross-regional spatiotemporal reasoning~\cite{gurnee2024languagemodelsrepresentspace}. These evaluations demonstrate that LLMs are capable of handling broad geographic contexts and structured facts. However, this focus on macro-scale capabilities obscures a critical limitation examined in this work: \textit{LLMs continue to struggle with hyper-local knowledge}, the fine-grained, community-specific information essential for grounded, real-world applications that demand geographic and cultural nuance.

The demand for hyper-local AI capabilities is rapidly growing. Civic platforms rely on AI systems that can navigate local governance structures and community dynamics~\cite{guridi2025thoughtful}, while community journalism initiatives need AI tools that contextualize events within local cultural narratives~\cite{ai_local_news}. Location-aware services must reason about neighborhood-specific preferences and constraints~\cite{chen2024travelagent, tang2024itinera,gao2025clips}. Despite these needs, current LLMs often default to generic outputs or propagate biases, failing to capture \textit{the multifaceted nature of local knowledge}, which includes not only statistical facts but also cultural norms, vernacular expressions, and community interactions, etc.~\cite{gao2025aturing}.

Existing benchmarks have made important progress in evaluating geographic knowledge, but they fall short in capturing the fine-grained complexity of local reasoning required for real-world, community-centered applications. Prior work focuses largely on global factual recall~\cite{moayeri2024worldbench, almeida2025tiebe}, isolated cultural references~\cite{shi2025libra, dudy2025unequal}, or generative descriptions of place identity in urban settings~\cite{jang2024place}, often targeting specific regions, subjective narratives, or coarse-grained indicators. Other efforts highlight geographic biases in model performance~\cite{manvi2024geobias, zhu2024quite}, or develop tools for geospatial inference~\cite{manvi2023geollm, wu2024torchspatial}, but do not offer comprehensive frameworks for local knowledge reasoning and evaluation.
A core reason for this gap is that existing evaluations lack the geographic granularity and knowledge breadth needed to capture how local knowledge operates in practice. Local reasoning is not just about factual recall or cultural trivia, but requires integrating statistical indicators, cultural narratives, vernacular expressions, and community-specific governance knowledge across diverse social contexts \cite{gao_journeying_2024,gao2025clips}.

Focusing on \textit{county-level knowledge} offers a tractable yet underexplored path toward evaluating local reasoning: (i) counties are the smallest U.S. administrative unit with consistently reported socio-economic statistics (e.g., ACS, CDC, USDA), (ii) they map cleanly onto electoral and governance structures that drive civic decision-making, making it a practical sweet-spot between granularity and data availability. 
However, current LLMs systematically overrepresent major metropolitan areas, leaving rural and smaller communities neglected~\cite{manvi2024geobias, sun2023head}. Addressing this imbalance requires an evaluation framework that reflects the full spectrum of localities, especially those that are underrepresented due to data scarcity. Moreover, it remains unclear whether LLMs can overcome these limitations through web search augmentation, particularly when community-specific knowledge is fragmented or absent from standard retrieval sources.

In this work, we introduce \textsc{LocalBench}, a benchmark for evaluating LLMs on county-level local knowledge and reasoning. Grounded in the Localness Conceptual Framework~\cite{gao2025aturing}, which defines local knowledge across physical, cognitive, and relational dimensions, \textsc{LocalBench} comprises 14,782 question–answer pairs across 526 U.S. counties. It spans all dimensions of the framework and is validated via expert annotation.

Using this benchmark, we evaluate 13 leading LLMs under both closed-book and web-augmented settings. Our findings reveal clear limitations: the best-performing models achieve only 56.8\% accuracy on narrative-style questions and struggle with numerical reasoning, falling below 15.5\%. Furthermore, increased model scale and current implementations of web augmentation do not guarantee better performance: web search improves Gemini (+13.6\%) but harms GPT-series models (–11.4\%), and larger or mixture-of-experts (MoE) architectures show no consistent advantage over smaller non-MoE models.

Our contributions are:
\begin{itemize}
\item \textbf{\textsc{LocalBench}:}
We introduce the first benchmark targeting U.S.\ county-level local knowledge and reasoning to advance research in place-aware AI. It contains 14,782 QA pairs covering 526 counties, spans all dimensions of localness, and is validated through expert annotation.
\item \textbf{Comprehensive LLM evaluation:}
We present a systematic evaluation of 13 state-of-the-art LLMs in both closed-book and web-augmented settings that reveals how even today’s strongest models falter when confronted with fine-grained, place-aware queries.
\item \textbf{Empirical insights for place-aware LLMs:}  
We provide evidence that challenges the common assumption that increased model size or retrieval augmentation inherently improves local reasoning, offering actionable insights for developing future place-aware language models.
\end{itemize}




\section{Related Work}

\subsection{Localness and Community-Centered Knowledge}

Digital-placemaking research shows that community platforms (e.g., neighborhood forums, local subreddits) rely on hyper-local information, such as vernacular expressions, governance details, and shared narratives, to foster civic engagement~\cite{aubin2024not, park2014understanding, gao_journeying_2024,gao2025collective}. LLMs are now used in this space, yet studies reveal that their place descriptions are often generic or stereotyped, especially outside iconic cities~\cite{jang2024place, zhu2024quite}. Existing benchmarks probe single facets: global facts \cite{moayeri2024worldbench}, regional term recognition~\cite{shi2025libra}, or city stereotypes~\cite{jang2024place}, but none test whether models grasp the full texture of locality.

To fill this gap, we adopt the Localness Conceptual Framework \cite{gao2025aturing}, which organizes local knowledge into three interwoven domains: Physical (direct interaction with a place), Cognitive (cultural and local knowledge), and Relational (social connections and emotional bonds), covering 7 dimensions and 88 subcomponents. This structure aligns with ``sense of place'' theory \cite{lengen2012sense} and provides a conceptual definition that allows us to ask whether an LLM can, for example, (i) recognize patterns of long-term residence (Physical), (ii) recount local history (Cognitive), and (iii) describe forms of community engagement (Relational).

By using localness as our evaluation lens we move beyond generalized location-focused knowledge like ``The Statue of Liberty is in New York City,'' and evaluate whether AI systems can surface hyper-local narratives, vernacular, and community dynamics across urban, suburban, and rural contexts. In short, we explore if LLMs and agentic web search approaches are sufficiently capable to be trusted for community-focused applications.

\subsection{Geographic Knowledge and Cultural Locality Benchmarks}

Existing benchmarks have made progress in evaluating LLMs’ geographic knowledge, but they often focus on global or national-level factual recall. 
\emph{WorldBench}~\cite{moayeri2024worldbench} tests LLMs on factual recall of national indicators such as GDP and literacy rates across over 180 countries. \emph{TiEBe}~\cite{almeida2025tiebe} evaluates temporal event recall at global and regional scales, focusing on historically significant world events. These benchmarks prioritize breadth over locality, without addressing fine-grained community-specific knowledge.

Other efforts target cultural locality or factual transfer in regional contexts, but often with limited geographic scope or narrow task design.
\emph{LIBRA}~\cite{shi2025libra} focuses on New Zealand-specific cultural terms, probing local bias through term recognition and classification. \emph{LoFTI}~\cite{dudy2025unequal} examines factuality transfer by prompting LLMs to adapt general knowledge into Indian state- and city-level contexts. \emph{Place Identity}~\cite{jang2024place} evaluates generative descriptions of cities, but focuses exclusively on urban areas, leaving suburban and rural communities unexamined.

These benchmarks address valuable aspects of cultural knowledge but lack systematic evaluation of local knowledge across diverse geographic and social settings.

\subsection{Geospatial Reasoning and Spatial Bias Probing}
Other benchmarks target spatial reasoning and geographic bias, but do not comprehensively evaluate local knowledge reasoning.
\emph{TorchSpatial}~\cite{wu2024torchspatial} assesses geospatial representation learning using geo-tagged image classification tasks, while \emph{GeoLLM}~\cite{manvi2023geollm} probes population inference and location classification via OpenStreetMap and WorldPop data. These efforts focus on spatial embeddings and regression tasks, rather than reasoning over local cultural or governance contexts. 

Complementing these tasks, several studies highlight geographic performance biases in LLMs, particularly their tendency to perform better on well-documented, high-resource regions while struggling with underrepresented communities~\cite{manvi2024geobias,zhu2024quite}. These works document systematic failures in representing less-popular or under-resourced areas, but focus primarily on analyzing model outputs across regions rather than providing structured benchmarks for evaluating local knowledge breadth and depth.

Broader efforts in geoscience and geospatial foundation models address related challenges but remain orthogonal to the problem of local knowledge reasoning. \emph{GeoGPT}~\cite{zhang2023geogpt} introduces a tool-augmented pipeline for executing geospatial tasks and querying geospatial APIs, while \emph{Contrastive Spatial Pretraining (CSP)}~\cite{mai2023csp} develops multimodal foundation models via visual-text contrastive learning on remote sensing data. These methods advance spatial understanding but do not target natural language evaluation of local, community-level knowledge.




\begin{figure*}[t]
    \centering
    \includegraphics[width=0.80\linewidth]{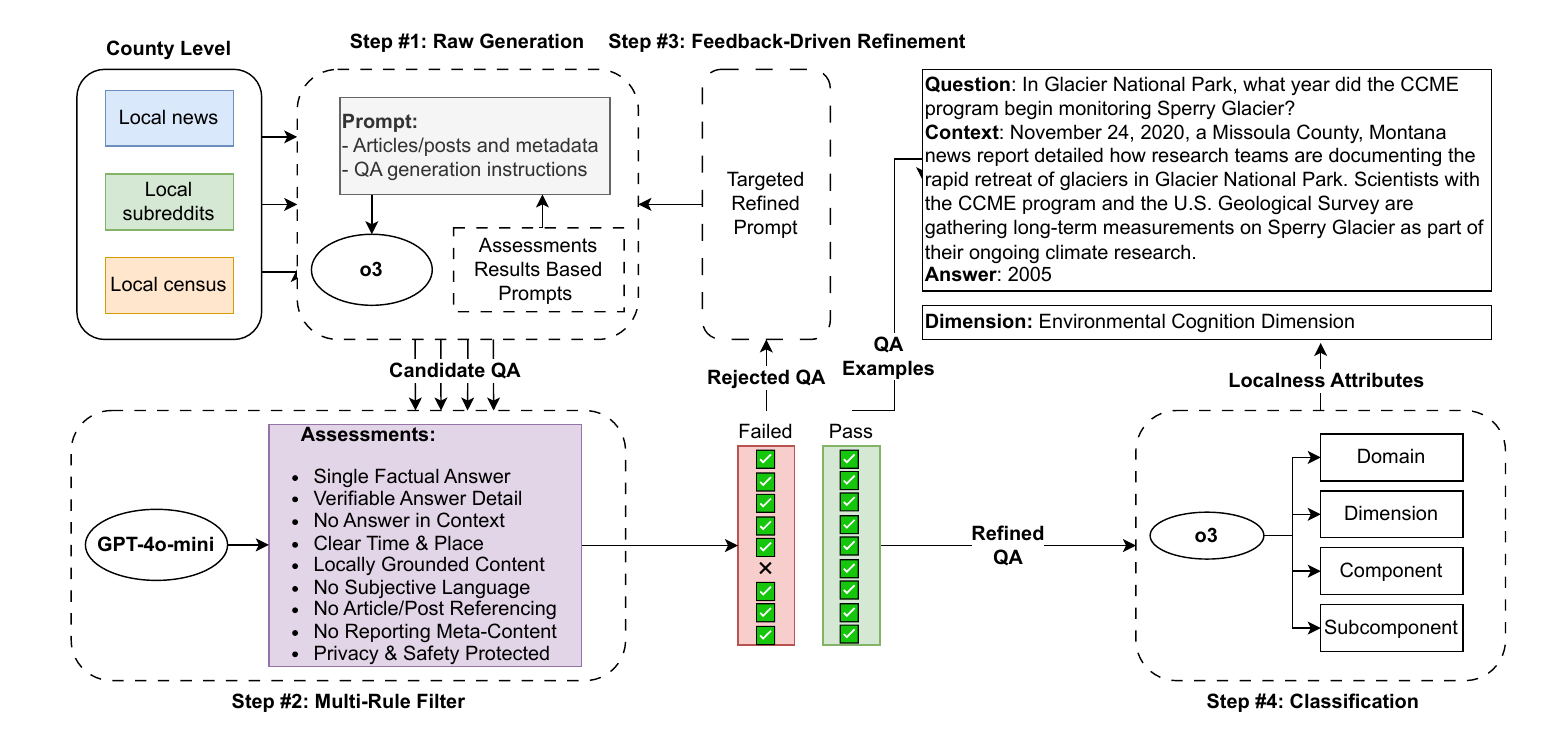}
    \caption{\textsc{LocalBench} construction pipeline. The process involves QA generation using a reasoning model and quality analysis via LLM-based assessment and filtering.}
    \label{fig:qa_pipeline}
\end{figure*}

\section{Benchmark Construction}
\label{sec:benchmark}

We introduce \textsc{LocalBench}, a large-scale benchmark designed to evaluate LLMs' ability to reason over county-level local knowledge. The dataset comprises 14,782 question–answer (QA) pairs covering 526 U.S. counties, with a balanced distribution across urban, suburban, and rural regions. Each QA pair is aligned to a ground-truth source and annotated according to the \emph{Localness Conceptual Framework}.

\subsection{Nature of Localness Evaluation}

\textsc{LocalBench} targets a core limitation of current LLMs: reasoning about hyper-local knowledge that is often difficult to retrieve online. Many questions require information from small local media outlets such as county newsletters, or community-level discussions from hyper-local forums like Reddit threads. Other questions demand understanding of fragmented institutional data, such as neighborhood-specific census metrics or municipal reports. These sources are rarely indexed comprehensively by search engines, are inconsistently formatted, and frequently require interpretive aggregation rather than direct retrieval \cite{gao_journeying_2024}.

\subsection{Data Sources}

\textsc{LocalBench} integrates three complementary data sources that together capture both structured and unstructured local knowledge. From the U.S. Census Bureau and other census data sources like USDA Agricultural Statistics Service, National Register of Historic Places, we extract 34 localness indicators, spanning cognitive, physical, and relational domains as described by \citet{gao2025aturing}. Examples include the cropland fertilization rate (cognitive), the percentage of residents living in the same home for more than five years (physical), and total ballots cast in the 2020 election (relational). 

Using Rural-Urban Continuum Codes (RUCC), we stratified counties into urban (RUCC 1–3), suburban (4–6), and rural (7–9) groups. From the 681 counties with complete data across all 34 indicators, we sample 60 counties per group (N=180). This structured source yields 6,120 QA pairs directly incorporated into the final dataset without quality filtering, evenly split between numerical questions (e.g., ``What is the median household income in County X?'') and comparison questions (e.g., ``How does County X's unemployment rate compare to County Y's?''). 
Appendix 14 lists the corresponding QA pairs.

To capture unstructured local discourse, we further collect subreddit data from ``The Global List of Local Reddits,'' from r/LocationReddit. We then manually inspected each subreddit and associated each subreddit with its corresponding county. Data from January 2024 to March 2025 includes posts and the top-50 comments per thread, initially generating 4,210 candidate QA pairs, yielding 4,000 final QA pairs after quality filtering. These final QA pairs focus on narrative and interpretive aspects of local culture, events, and community concerns.

Additionally, we use the NELA-Local corpus~\cite{horne2022nela}, containing over 1.4 million local news articles from 313 U.S.\ outlets published between April 2020 and December 2021. Articles are tagged at the county level, initially generating 4,897 candidate QA pairs, which results in 4,662 final QA pairs after quality filtering. These final QA pairs cover governance, civic activities, and hyper-local events that were reported in local news outlets.

Across all sources, the pipeline processes 15,527 initial generation attempts, achieving an overall success rate of 96.9\% to produce the final dataset of 14,782 QA pairs covering 526 unique counties across 49 states, with geographic diversity confirmed via Moran's $I = -0.003$ ($p=0.491$), indicating no spatial autocorrelation.

\subsection{QA Generation Pipeline}

\subsubsection{Step \#1: Raw Generation}
Figure~\ref{fig:qa_pipeline} illustrates the three-stage pipeline for QA pair generation and validation. In the first stage, we use the OpenAI o3 model to generate candidate QA pairs from the source materials. The generator operates with a temperature parameter of 0.7, top-p parameter of 0.9, and max\_tokens parameter of 200, producing 1–3 QAs per document to balance diversity and quality. Importantly, generation is constrained to reason over the given input rather than hallucinate external knowledge, ensuring county-specific grounding. 

\subsubsection{Step \#2: Multi-Rule Filter}
Census-derived QA pairs bypass the quality control pipeline due to their structured nature and verified accuracy. For Reddit and news sources (9,107 candidate pairs), a QA Quality Analyzer filters the generated pairs using a fine-tuned GPT-4o-mini model trained via Direct Preference Optimization (DPO)~\cite{rafailov2023direct}. The training dataset consists of 473 human-annotated QA pairs, 
which were labeled by two graduate researchers with inter-annotator agreement $\kappa=0.84$. The analyzer evaluates each pair on nine criteria, including single factual answer validation, geographic grounding, subjectivity filtering, privacy preservation, and temporal consistency, and others (detailed in 
Appendix 3). Training parameters include a learning rate of one, batch size of eight, and two epochs. 

\subsubsection{Step \#3: Feedback-Driven Refinement}
Reddit and news-derived pairs failing any quality criterion enter an iterative re-generation loop, where the QA Generator receives targeted feedback and retries up to three times. If a pair fails all attempts, we remove it from the dataset. After three regeneration rounds, we retain 95.2\% of non-census pairs, contributing to the overall 96.9\% dataset success rate.

\textbf{\emph{Human Verification}.} To validate filter accuracy, two independent human annotators annotated 500 randomly sampled QA pairs. The annotators achieved inter-annotator agreement of $\kappa=0.78$. When comparing with our QA Quality Filters' output, we observed an overall F1-score of 0.94. There is no meaningful correlation across error cases, suggesting that our QA Quality Analyzer produces QA pairs of equivalent quality across different kinds of QA pairs (correlation 
analysis results are in Appendix 5).

\subsubsection{Step \#4: Localness Attribute Classification}
Each validated QA pair undergoes localness classification according to the Localness Conceptual Framework using the o3 model. The classifier receives the question, answer, and original source context, then assigns labels across four hierarchical levels: domain (Physical/Cognitive/Relational), dimension, component, and subcomponent. Classification operates with a temperature parameter of 0.3 and a max\_tokens parameter of 150 to ensure consistent, focused outputs. The classification prompt instructs the model to analyze the QA pair's core focus and assign the most specific applicable labels. This hierarchical structure ensures comprehensive coverage while maintaining classification granularity needed for detailed analysis.

\textbf{\emph{Human Verification}.} 
To validate classification accuracy, two independent human annotators annotated 200 randomly sampled QA pairs. The annotators achieved inter-annotator agreement of $\kappa=0.87$. When comparing human annotations against o3 classifications, we observed 94.2\% precision across all classification levels, with domain-level precision reaching 98.5\%. 

\subsection{Dataset Statistics}
\label{sec:dataset_stats} 

Table~\ref{tab:dataset_statistics} provides comprehensive statistics for the final \textsc{LocalBench} dataset, broken down by localness components and data sources. The dataset exhibits balanced coverage across all major localness categories while maintaining consistent content complexity across components.

\begin{table*}
\centering
\small
\resizebox{0.95\textwidth}{!}{%
\begin{tabular}{@{}lccccccccccc@{}}
\toprule
\textbf{Domain} & \textbf{Dimension} & \textbf{Number} & \textbf{QLen} & \textbf{CLen} & \textbf{ALen} & \textbf{Census} & \textbf{Reddit} & \textbf{News} & \textbf{Rural} & \textbf{Suburban} & \textbf{Urban} \\
\midrule
\multirow{2}{*}{\textbf{Physical}} 
& Place Interaction & 1,330 & 28.9 & 63.9 & 5.3 & 180 & 663 & 487 & 306 & 337 & 687 \\
& Temporal Presence & 2,907 & 29.2 & 62.4 & 4.2 & 1260 & 567 & 1,080 & 724 & 818 & 1,365 \\
\midrule
\multirow{3}{*}{\textbf{Cognitive}} 
& \makecell[c]{Cultural\\Understanding} & 1,435 & 28.5 & 63.6 & 4.4 & 720 & 286 & 429 & 352 & 411 & 672 \\
& Environmental Cognition & 1,739 & 30.3 & 63.2 & 5.3 & 540 & 860 & 339 & 519 & 471 & 749 \\
& Local Knowledge & 3,855 & 28.5 & 63.1 & 4.5 & 1620 & 914 & 1,321 & 987 & 1,088 & 1,780 \\
\midrule
\multirow{2}{*}{\textbf{Relational}} 
& Emotional Connection & 838 & 26.8 & 64.2 & 3.0 & 720 & 54 & 64 & 254 & 267 & 317 \\
& \makecell[c]{Social/Community \\ Engagement} & 2,678 & 27.5 & 63.6 & 4.5 & 1080 & 656 & 942 & 651 & 774 & 1,253 \\
\midrule
\makecell[c]{\textbf{Overall}} & -- & \textbf{14,782} & \textbf{28.49} & \textbf{63.45} & \textbf{4.43} & \textbf{6,120} & \textbf{4,000} & \textbf{4,662} & \textbf{3,793} & \textbf{4,166} & \textbf{6,823} \\
\bottomrule
\end{tabular}
}
\captionsetup{width=.95\textwidth,singlelinecheck=false}
\caption{Dataset statistics by localness dimensions and data sources, with counts by RUCC group. \textbf{QLen}, \textbf{CLen}, and \textbf{ALen} represent the average token lengths of the \textit{Query}, \textit{Context}, and \textit{Answer}, respectively. Token counts are measured using the GPT-4o tokenizer.}
\label{tab:dataset_statistics}
\end{table*}

\section{Evaluation Setup}
\label{sec:evaluation}

\subsection{Evaluation Protocol}

As noted above, our analysis focuses on both prominent proprietary and open source LLMs, both without web search augmentation, and where possible, with web search augmentation as well. To ensure reproducibility, all models are run with temperature parameters set to 0.0, and maximum token output parameters of 256 tokens. Each model receives standardized prompts designed to elicit factual, grounded responses while allowing models to express uncertainty when appropriate. Each model is evaluated over three independent runs with different random seeds and we report mean scores.
For statistical comparisons, we conduct paired t-tests with Bonferroni correction ($\alpha=0.05$) to control for multiple comparisons.

\subsection{Models Evaluated}

We evaluate a diverse set of models across different capability tiers. Proprietary models include GPT-4o (gpt-4o-2024-08-06), GPT-4.1 (gpt-4.1-2025-04-14)~\cite{gpt}, Gemini-2.5-Pro, Gemini-2.5-Flash~\cite{gemini}, Claude-4-Sonnet (claude-sonnet-4-20250514), Claude-3.7-Sonnet (claude-3-7-sonnet-20250219)~\cite{claude}, Qwen3-235B-A22B/30B-A3B/32B/14B/8B~\cite{yang2025qwen3}. We also test web-augmented configurations, including GPT-4.1 with Search API integration and Gemini-2.5-Pro with Search Grounding.

\subsection{Evaluation Metrics}

We adopt a multi-faceted evaluation framework designed to capture factual correctness, semantic equivalence, numeric reasoning accuracy, and model confidence alignment. This is necessary to address the diverse formats and ambiguity present in locality-grounded questions. In all cases, we pose the question in our QA pairs to the LLM, and collect its generated answer. We then compare the generated answer to our ground-truth answer in our QA pair, based on these metrics. 

To evaluate factual correctness, we compute both \textbf{Exact Match (EM)}, which enforces strict string equality after normalization, and \textbf{ROUGE-1 F1}, which tolerates paraphrasing through unigram overlap. To capture deeper equivalence beyond lexical variation, we compute a \textbf{semantic match} score: the cosine similarity between dense embeddings of the generated answer and our ground-truth answer, obtained with OpenAI’s \texttt{text-embedding-3-small}.

For numerical answers, we report numerical \textbf{accuracy}, which counts a numerical prediction as correct if its relative error is under 2\% of the gold value (and requires an exact zero when the gold answer is zero).

To handle ambiguous cases where answers may vary in surface form but remain valid, we introduce a \textbf{GPT Judge} mechanism. A GPT-4o-mini model receives the full QA context: original question, gold answer, generated answer, and any supporting evidence, and outputs a binary judgment of correctness. We also capture the log-probability of the generated judgment token (e.g., ``Correct'' or ``Incorrect''), allowing us to interpret the model's confidence in its assessment. To validate our GPT-based judge, we conduct human annotation on 200 randomly selected samples. The results show that the judge aligns with human annotations in 96\% of cases, which demonstrates its reliability for downstream evaluation.

Lastly, we compute the \textbf{answer rate} as the proportion of model outputs that provide substantive responses (i.e., excluding empty or ``I don't know'' answers), reflecting model willingness to engage with local queries.




\begin{table*}[ht]
\centering
\label{tab:llm_mean_performance}
\small
\begin{tabular}{l|ccccc|cc}
\toprule
\multirow{2}{*}{\textbf{Model}} & \multicolumn{5}{c|}{\textbf{Non-Numerical QA}} & \multicolumn{2}{c}{\textbf{Numerical QA}} \\
& \textbf{EM} & \textbf{ROUGE-1} & \textbf{Semantic} & \textbf{GPT Judge} & \textbf{Ans Rate} & \textbf{Accuracy} & \textbf{Ans Rate} \\
\midrule
GPT-4o & 22.0 & 30.7 & 53.0 & 32.8 & 99.6 & 6.2 & 39.8 \\
GPT-4.1 & \textbf{32.2} & \textbf{52.5} & \textbf{74.1} & 47.0 & 100.0 & 6.2 & 100.0 \\
GPT-4.1+Web & 13.5 & 27.9 & 43.2 & 35.6 & 92.9 & \textbf{15.5} & 92.0 \\
\midrule
Gemini-2.5-Pro & 28.0 & 52.0 & 70.5 & 52.5 & 100.0 & 12.8 & 100.0 \\
Gemini-2.5-Flash & 31.1 & 46.0 & 67.6 & 43.2 & 100.0 & 7.5 & 100.0 \\
Gemini-2.5-Pro+Grounding & 21.9 & 50.1 & 66.0 & \textbf{56.8} & 91.7 & 12.8 & 100.0 \\
\midrule
Claude-Sonnet-4 & 23.4 & 38.5 & 64.0 & 39.7 & 100.0 & 7.1 & 97.3 \\
Claude-Sonnet-3.7 & 21.7 & 42.5 & 65.5 & 43.7 & 100.0 & 8.4 & 91.2 \\
\midrule
Qwen3-235B-A22B & 19.9 & 29.0 & 54.0 & 27.3 & 99.3 & 6.6 & 77.0 \\
Qwen3-30B-A3B & 20.5 & 29.6 & 54.9 & 28.0 & 99.7 & 2.2 & 100.0 \\
Qwen3-32B & 20.0 & 29.9 & 55.4 & 27.7 & 99.7 & 4.9 & 99.1 \\
Qwen3-14B & 19.5 & 29.8 & 55.4 & 27.5 & 99.6 & 4.0 & 100.0 \\
Qwen3-8B & 16.3 & 27.2 & 54.1 & 22.9 & 99.6 & 3.1 & 75.2 \\
\bottomrule
\end{tabular}
\caption{Performance Results Across Non-Numerical and Numerical QA}
\end{table*}

\section{Results}
\label{sec:results}

We evaluate the performance of state-of-the-art LLMs on \textsc{LocalBench} to assess their capacity for local knowledge reasoning. Our analysis reveals substantial challenges that persist across task types, model architectures, and augmentation strategies.

A major finding is the sharp divide between non-numerical and numerical performance. While models such as \textit{Gemini-2.5-Pro+Grounding} reach up to 56.8\% GPT Judge accuracy on non-numerical questions, performance on numerical Census tasks remains critically low, with no model exceeding 12.8\% accuracy. We observe that models sometimes refuse to answer numerical queries by explicitly stating a lack of knowledge. In particular, \textit{GPT-4o} answers only 39.8\% of such questions, whereas all other later-released models respond to at least 75\%. This discrepancy may reflect updates in training data or changes in post-training strategies over time, amid the rapidly evolving landscape of LLM development.
Web augmentation further exposes architectural differences in retrieval integration. Gemini models benefit substantially from external evidence, while surprisingly, \textit{GPT-4.1+Web} exhibits degraded performance and reduced answer rates, suggesting that retrieval-based grounding can increase model uncertainty when poorly integrated.
Model scaling does not consistently improve local knowledge reasoning. Larger models often underperform their smaller counterparts, and mixture-of-experts architectures fail to outperform dense models. In particular, spatially grounded numerical reasoning remains elusive even for the largest proprietary models, indicating that these limitations stem not from scale but from representational and architectural mismatches.

Taken together, these results highlight the persistent difficulty of local knowledge tasks for current LLMs. Addressing these limitations will require new modeling approaches that explicitly encode geographic reasoning and better handle context-dependent, place-specific information.

\subsection{Breaking Down Model Performance}

While average accuracy offers a basic ranking of models, effect size analysis reveals the magnitude and reliability of performance differences.
We quantify performance differences with \emph{Cohen's $d$}, a standardized mean‐difference that expresses how many pooled standard deviations separate two systems; values near~$0.20$,~$0.50$, and~$0.80$ are conventionally interpreted as small, medium, and large effects, respectively~\cite{cohen2016power}.  
To protect the family‐wise error rate during the many pairwise contrasts, we adjust each $p$‐value with the Bonferroni procedure~\citep{bland1995multiple}  
(results details in Appendix 9 and 10).

For non-numerical tasks, the \textit{Gemini-2.5-Pro+Grounding} achieves the highest performance with 56.8\% accuracy and a large positive effect size ($d = +0.485, p<0.001$) relative to \textit{GPT-4.1}. \textit{GPT-4o} shows significant performance degradation with 32.8\% accuracy and a large negative effect size ($d = -0.412, p<0.001$). Open-source models lag substantially with large negative effect sizes (Qwen3-30B: $d = -0.581, p<0.001$).

However, for numerical  tasks, overall performance remains critically low across all models. The highest-performing models (\textit{GPT-4.1+Web}) achieve only 15.5\% accuracy, representing a substantial but still modest improvement ($d = +0.623, p<0.001$) over the 6.2\% baseline. Most concerning is the dramatic variation in answer rates, while most models maintain answer rates greater than 95\%, \textit{GPT-4o} drops to 39.8\%, indicating systematic avoidance of numerical reasoning. Notably, open-source models exhibit broadly similar numerical reasoning capability to \textit{GPT-4o}.

\subsection{Web-Augmented Model Behavior}

We find that web augmentation produces opposing effects across model families. For non-numerical QA, \textit{Gemini-2.5-Pro+Grounding} shows substantial improvements (+13.6\% in GPT Judge accuracy), achieving the benchmark's highest performance. In stark contrast, \textit{GPT-4.1+Web} shows performance degradation (-11.4\%), suggesting fundamental differences in retrieval integration capabilities. Further, we see that patterns in model answer rate suggest retrieval confidence issues. \textit{GPT-4.1+Web} and \textit{Gemini-2.5-Pro+Grounding} show reduced willingness to answer from 100.0\% to 92.9\% and 91.7\%, respectively, indicating that retrieved information may increase uncertainty in the model, somewhat counterintuitively. 
For numerical QA, \textit{GPT-4.1+Web} and \textit{Gemini-2.5-Pro+Grounding} both show substantial improvements (+9.3\% and +5.3\%), while \textit{GPT-4.1+Web} shows reduced willingness to answer from 100.0\% to 92.0\%.
Our results indicate that web augmentation effectiveness is highly model-dependent, with architectural differences in retrieval processing creating divergent outcomes for local knowledge reasoning tasks, depending on the model being used.

\subsection{Scaling Analysis}
\label{sec:scaling}

We find that local knowledge reasoning shows limited or even negative scaling effects. Within the Qwen family of models, performance gains taper off beyond 32B parameters and even regress at larger scales. This breakdown is particularly evident when comparing mixture-of-experts (MoE) and non-MoE models. Within the Qwen series, the 235B MoE model underperforms not only its distilled 30B counterpart but also the smaller 32B and 14B non-MoE models. These results suggest that MoE architectures may struggle to encode spatial and contextual nuances necessary for local knowledge.

We also find that the limitations of scaling are most acute for numerical local knowledge. Even state-of-the-art proprietary models achieve near-zero accuracy on such questions ranging from 2.2\% to 15.5\%. This suggests a qualitative limitation in model design rather than a lack of training data or parameter count: current models appear fundamentally constrained in their ability to represent place-specific quantitative reasoning. Further, these results suggest that local knowledge reasoning is not constrained by model size itself, but by mismatches between transformer architectures and the structured, spatially grounded nature of local information. Explicit geographic and place-aware reasoning and representations are important directions for future work.

\section{Discussion}
\label{sec:discussion}

Our evaluation reveals that local knowledge reasoning represents a qualitatively different challenge for current LLMs. The systematic breakdown across numerical reasoning, cultural contextualization, and scaling relationships suggests fundamental architectural limitations rather than simple knowledge gaps.

\subsection{The Architectural Mismatch Problem}

The failure of numerical reasoning may indicate deeper issues than ``mere'' memory limitations. Current transformer architectures, optimized for next-token prediction on global text corpora, appear fundamentally misaligned with the precision required for quantitative local reasoning. This mismatch is compounded by the situated nature of local knowledge, where factual claims depend on geographic, temporal, cultural, and ultimately human context. Current architectures lack explicit mechanisms for spatial reasoning, temporal grounding, or cultural contextualization, which are capabilities essential for local knowledge understanding. The MoE routing failures further demonstrate that scaling expert capacity does not, today, compensate for these representational gaps.

\subsection{The Retrieval Integration Paradox}

The opposing effects of web augmentation across model families reveal that retrieval integration depends heavily on underlying architectural capabilities rather than search quality alone. The unstable augmentation effectiveness suggests differential capacity for filtering noisy content and maintaining uncertainty calibration when external information conflicts with parametric knowledge. This set of results challenges the prevailing assumption that better search algorithms will help solve grounding problems. Instead, it may be that the challenge lies in developing architectures that can effectively discriminate between high- and low-quality local sources of information and synthesize information across heterogeneous community knowledge systems.

\subsection{Implications for Future Work}

These findings argue for fundamental shifts in AI development priorities. Rather than pursuing larger models trained on comprehensive web corpora, the field must invest in architectures specifically designed for situated reasoning, including spatial reasoning capabilities, cultural contextualization mechanisms, and uncertainty estimation methods that handle local knowledge ambiguity. Importantly, the breakdown of conventional scaling laws for local knowledge tasks indicates that current paradigms cannot address place-based reasoning through scale alone. This may necessitate architectural innovations, including geographic reasoning modules and place-aware attention mechanisms.

\subsection{Limitations}
\label{sec:limitations}

\textsc{LocalBench} is constructed from data spanning 2020 to 2025 across 526 U.S. counties. While this scope enables systematic benchmarking within a well-documented administrative structure, it also introduces important constraints. Most notably, the benchmark reflects only U.S.-based geographic, cultural, and institutional knowledge. As such, its findings may not generalize to local contexts in other countries, where governance structures, data availability, and sociolinguistic norms differ significantly. Extending local knowledge benchmarks internationally will require adapting both the conceptual framework and data sourcing strategies to diverse geopolitical settings.

Within the U.S., uneven digital footprints persist. Rural counties with limited online presence remain underrepresented~\cite{thebault-spieker_distance_2018, thebault-spieker_geographic_2018, johnson_not_2016}, mirroring broader structural disparities that \textsc{LocalBench} cannot fully resolve. In addition, our exclusive reliance on English-language sources restricts coverage of multilingual communities~\cite{hickman_understanding_2021}, including Indigenous, immigrant, and border populations whose local knowledge may be expressed in Spanish, Native languages, or other vernaculars not captured here.

While efforts were made to ensure data quality, some QA pairs may still contain factual inaccuracies or temporal mismatches, especially in regions with sparse digital activity. Although our evaluation framework integrates multiple metrics and GPT-based judgment (validated against human annotators), it may still miss nuances of cultural appropriateness or locally grounded truth, suggesting the need for human-in-the-loop approaches in future evaluations.

\section{Conclusion}
\label{sec:conclusion}

\textsc{LocalBench} demonstrates that local knowledge reasoning poses a fundamental challenge for current LLMs: one that cannot be solved by scale or retrieval alone. Our evaluation across 14,782 QA pairs and 526 U.S. counties uncovers persistent failures in numerical reasoning, cultural understanding, and confidence calibration, revealing architectural and representational mismatches. These limitations point to the need for spatially grounded models, better uncertainty handling, and participatory development practices that center community epistemologies. The observed gap between model confidence and accuracy is especially salient given prevailing conversations about the benefits and societal importance of AI, and our results suggest this is not true for all communities. While \textsc{LocalBench} offers a foundation for diagnosing these gaps, advancing equitable place-aware AI will require new architectures, ethical frameworks, and sustained collaboration between researchers, communities, and policymakers.


\section{Ethical Statement}

Local knowledge AI raises critical questions about community data sovereignty and representation. The situated nature of local knowledge, embedded in community relationships and cultural contexts, demands new approaches to responsible AI development beyond traditional ethics frameworks.

\textbf{Community Data Sovereignty:} Local knowledge often embodies community-shared experiences and interpretations that carry significant cultural value. Extraction of such knowledge for AI training without explicit community consent risks epistemic appropriation --- the commodification of community wisdom without acknowledgment or compensation. This concern is particularly acute for Reddit-derived data, where community members share information within specific social contexts inappropriate for external AI reuse.

\textbf{Participatory Development:} We advocate for community-centered AI development that treats local knowledge holders as partners rather than data sources, including community advisory boards for dataset development, benefit-sharing models, and opt-in governance mechanisms respecting community autonomy over knowledge sharing.




\bibliography{aaai2026}





\appendix
\onecolumn







\section{Appendix 1: Reddit Data QA Generation Prompt}

\begin{lstlisting}

You are given a single Reddit thread (post + comments) from a *local* subreddit of {county, state}. 

## From it, you will generate ONE QA pair that is:
- Clear, locally grounded about {county, state}, and answerable with one correct response.
- Either:
-- Track A: Fact-based Local QA (grounded in verifiable, non-subjective local information)
-- Track B: Community Insight QA (reflects clearly shared local community experiences)

## STRICT QA RULES: 
1. **NO THREAD-REFERENTIAL QUESTIONS.**  
- The QA pair should stand alone without relying on Reddit as a quoted source of authority or opinion.
- Forbidden phrases include: ''was mentioned as...'', ''according to the thread/discussion/Reddit post/comment/commenter'', ''what did commenters think'', etc.  
- Litmus test: *Would the question still make perfect sense to someone who never sees the thread?* If not, reject it.
2. FACT-BASED ONLY: Question MUST has **exactly one correct, fact-based answer.
- Prefer using the post's original question if applicable.  
- Include time/place qualifiers for precision.
- It does **not** depend on user preferences, multiple viewpoints, or recommendations
3. AVOID all of these:
- Questions asking how people felt, or vague "why" questions.
- Any framing that implies subjectivity, speculation, or unconfirmed information.
3. CONTEXT must:
- Be 2 to 4 neutral sentences summarizing thread purpose.
- Mention exact date (or month), year, county, and state.
- Attribute information sources.
- Do **not** leak answer.
4. ANSWER must:
- Track A: one exact, verifiable fact (from post or multiple comments).
- Track B: a consensus summary if multiple users echoed same view.
5. Prefer Track A. Use Track B only if consensus is unambiguous.
6. No meta-Reddit content (karma, mods, usernames).  
7. Privacy & Safety: no doxxing, no private phone numbers.

## POST DETAILS:
Post Title: {post_title}
Post Information:
- Date: {date}
- County: {county}
- State: {state}
Post Content:
{post_content}
Comments:
{post_comments}

## OUTPUT FORMAT:
[PAIR1]
Question: <one clearly answerable question>
Context: <neutral summary with time and location>
Answer: <fact or consensus summary>
Selected Comments: <e.g., 3,7,12>
Pair_type: <fact|insight>

## EXAMPLES OF ACCEPTABLE QUESTIONS: 
- What city program was in contract phase in March 2024 in Broomfield, Colorado?
- Which consulting firm was hired in June 2025 to evaluate Athens' zoning?
- What fungus linked to bird feces raised public health concern in Honolulu?
- Bad questions that are not acceptable: What do residents want? What do recent discussions say? Which business do people like? What do some suggest?

## HARD STOPS. Do NOT generate a QA if:
- The answer is not clearly stated or verifiable
- No single correct answer can be derived
- Content is entirely opinion-based or speculative

\end{lstlisting}

\section{Appendix 2: News Data QA Generation Prompt}
\begin{lstlisting}

You are given a local news article. 

From it, you will generate ONE QA pair that is:
- Clear, locally grounded, and answerable with one correct response.
- Either:
-- Track A: Fact-based Local QA (grounded in verifiable, non-subjective local information)
-- Track B: Community Insight QA (reflects clearly shared local community experiences) 

## ARTICLE DETAILS:
Article Title: {metadata['title']}
Article Information:
- Date: {metadata['date']}
- County: {metadata['county']}
- State: {metadata['state']}
- Source: {metadata['source']}
- Article Content (Factual Sentences): {news_article}

## STRICT QA RULES: 
1. FACT-BASED ONLY: Question must have ONLY **one correct answer**, and the **only answer** is clearly supported by the article content. No subjectivity.
- Good Example: \"What program was paused in March 2024?\"
- NOT: \"What platform was recommended?\" or \"What did residents suggest?\"
- Include time/place qualifiers for precision.
2. AVOID these:
- Questions asking how people felt, what they recommended, or vague \why\ questions.
- Any framing that implies subjectivity, speculation, or unconfirmed information.
- Questions about the news article itself (e.g., \"What does the article discuss?\", \"according to the article\", etc.).
3. CONTEXT must:
- Be 2-4 neutral sentences summarizing article purpose.
- Mention exact date (or month/year), county, and state.
- Attribute information sources (\''The article reported...\'', \''Officials noted...\'').
- Do **not** leak answer.
4. ANSWER must:
- Track A: one exact, verifiable fact from the article.
- Track B: a consensus summary if multiple sources echoed same view.
5. Track A is **preferred**. Only use Track B if consensus is unambiguous.
6. No meta-content (sources, journalists, publication details).
7. Privacy & Safety: no doxxing, no private phone numbers.

## OUTPUT FORMAT: 
[PAIR],
Question: <one clearly answerable question>,
Context: <neutral summary with time and location>,
Answer: <fact or consensus summary>,
Selected Sentences: <e.g., 3, 7, 12>,
Pair_type: <fact|insight>,

## EXAMPLES OF ACCEPTABLE QUESTIONS: 
- What city program was in contract phase in March 2024 in Broomfield, Colorado?,
- Which consulting firm was hired in June 2025 to evaluate Athens' zoning?,
- What public health concern was identified in Honolulu in 2024?,
- NOT: What do residents want? What do recent discussions say? Which business do people like?,

## HARD STOPS. Do NOT generate a QA if:,
- The answer is not clearly stated or verifiable,
- No single correct answer can be derived,
- Content is entirely opinion-based or speculative,
\end{lstlisting}

\section{Appendix 3: Quality Assessment Criteria}
\label{appendix:quality_criteria}

The DPO-tuned quality analyzer evaluates generated QA pairs against the following nine criteria:

\begin{enumerate}
\item \textbf{Single Factual Answer}: The question has one clear, factual answer that can be verified from the source material. Avoids questions with multiple valid interpretations or subjective responses.

\item \textbf{Geographic Grounding}: The question and answer are specifically tied to the mentioned county. Generic questions that could apply to any location are rejected.

\item \textbf{Subjectivity Detection}: The question and answer avoid subjective language, personal opinions, or value judgments. Focuses on factual, verifiable information.

\item \textbf{Privacy Compliance}: No personal identifiers, private information, phone numbers, addresses, or individual names are included in the QA pair.

\item \textbf{Safety Compliance}: Content does not promote harmful activities, contain offensive language, or include sensitive political commentary that could cause harm.

\item \textbf{Temporal Consistency}: The question and answer are consistent with the time period of the source material. Avoids anachronistic references or outdated information presented as current.

\item \textbf{Difficulty Assessment}: The question is neither trivial (answerable with basic general knowledge) nor impossibly difficult (requiring highly specialized expertise). Appropriate for evaluating local knowledge reasoning.

\item \textbf{Question Clarity}: The question is clearly formulated, unambiguous, and can be understood without additional context beyond the county specification.

\item \textbf{Answer Completeness}: The answer adequately addresses the question with sufficient detail while remaining concise. Avoids incomplete or overly brief responses that don't fully answer the question.
\end{enumerate}

\section{Appendix 4: Filters Accuracy}
\label{appen: filter_accuracy}

Table~\ref{tab:quality_accuracy} show each quality filter's performance against human annotation.

\begin{table}[H]
\centering
\begin{tabular}{@{}lccc@{}}
\hline
\textbf{Task} & \textbf{Accuracy} & \textbf{Recall} & \textbf{F1} \\
\hline
Single Factual Answer & 94.2 & 96.1 & 0.95 \\
Geographic Grounding & 96.8 & 94.3 & 0.96 \\
Subjectivity Detection & 89.3 & 91.7 & 0.91 \\
Privacy Compliance & 98.1 & 99.2 & 0.99 \\
Safety Compliance & 97.4 & 98.6 & 0.98 \\
Temporal Consistency & 92.7 & 89.8 & 0.91 \\
Difficulty Assessment & 88.6 & 92.4 & 0.90 \\
Question Clarity & 93.5 & 95.1 & 0.94 \\
Answer Completeness & 90.8 & 93.6 & 0.92 \\
\hline
\textbf{Overall} & \textbf{93.4} & \textbf{94.5} & \textbf{0.94} \\
\hline
\end{tabular}
\caption{Quality filters performance against human annotation.}
\label{tab:quality_accuracy}
\end{table}

\section{Appendix 5: Task Correlation Analysis}

We computed pairwise correlations between quality task failures to quantify their independence. Table~\ref{tab:err_corr_table} shows the results.

\begin{table}[H]
\centering
\begin{tabular}{l|*{8}{c}}
\hline
\textbf{Task} & \textbf{SFA} & \textbf{GG} & \textbf{SD} & \textbf{PC} & \textbf{SC} & \textbf{TC} & \textbf{DA} & \textbf{QC} \\
\hline
Geographic Grounding & 0.12 & -- & & & & & & \\
Subjectivity Detection & 0.31 & 0.08 & -- & & & & & \\
Privacy Compliance & 0.02 & 0.01 & 0.05 & -- & & & & \\
Safety Compliance & 0.04 & 0.03 & 0.15 & 0.18 & -- & & & \\
Temporal Consistency & 0.18 & 0.22 & 0.09 & 0.01 & 0.02 & -- & & \\
Difficulty Assessment & 0.25 & 0.19 & 0.14 & 0.03 & 0.06 & 0.11 & -- & \\
Question Clarity & 0.28 & 0.16 & 0.20 & 0.05 & 0.07 & 0.13 & 0.33 & -- \\
Answer Completeness & 0.22 & 0.11 & 0.17 & 0.04 & 0.08 & 0.15 & 0.29 & 0.35 \\
\hline
\end{tabular}
\caption{Pearson correlation coefficients between quality task failures (SFA=Single Factual Answer, GG=Geographic Grounding, SD=Subjectivity Detection, PC=Privacy Compliance, SC=Safety Compliance, TC=Temporal Consistency, DA=Difficulty Assessment, QC=Question Clarity). Low correlations indicate task independence.}
\label{tab:err_corr_table}
\end{table}

The low correlation coefficients (mean $r = 0.14$, max $r = 0.35$) confirm that quality assessment tasks capture largely independent failure modes, justifying the multi-criteria approach.

\section{Appendix 6: GPT Judge Prompts}
\label{appendix:judge_prompts}

\begin{lstlisting}
Evaluate if the AI-generated answer is correct based on the question and golden answer.

- Question: {question}
- Context: {context}
- Golden Answer: {gold_answer}
- AI Answer: {pred_answer}

## Instructions:
- Answer "Yes" if the AI answer is factually correct and addresses the question
- Answer "No" if the AI answer is factually incorrect or doesn't address the question
- Consider partial credit for answers that are mostly correct
- Ignore minor wording differences if the core meaning is correct

Answer (Yes/No):
\end{lstlisting}

\section{Appendix 7: Iterative QA Re-generation Process}
\label{appendix:regeneration}

\subsection{Feedback-Driven Re-generation}

When QA pairs fail quality assessment, they are returned to the QA Generator with targeted feedback. The re-generation prompt incorporates specific guidance based on the failed criteria:

\subsubsection{Example Re-generation Feedback}

\textbf{Original Failed QA Pair:}
\begin{itemize}
\item Q: ``What do people think about the schools in Adams County?''
\item A: ``Most residents are satisfied with the educational quality.''
\item \textbf{Failed Criteria}: Single Factual Answer, Subjectivity Detection
\end{itemize}

\textbf{Re-generation Prompt:}
\begin{lstlisting}
\texttt{The previous QA pair failed quality assessment for the following reasons:
- Question allows multiple valid interpretations and subjective responses
- Answer contains opinion-based language without factual grounding

Please generate a new QA pair that:
1. Has a single, factual answer verifiable from the source
2. Avoids subjective language and opinion-based content
3. Focuses on measurable aspects of education in Adams County

Use the same source material: [SOURCE\_CONTENT]}
\end{lstlisting}

\subsection{Re-generation Success Analysis}

Analysis of re-generation patterns reveals that different failure types have varying recovery rates. 

Privacy and safety compliance issues show the highest recovery rates, as these typically involve simple content modifications or redaction. Conversely, difficulty assessment and subjectivity detection prove most challenging to address, often requiring fundamental reconceptualization of the question-answer relationship. Geographic grounding issues show moderate recovery rates, as they usually require restructuring questions to include county-specific context rather than generic phrasing.

\section{Appendix 8: Localness Annotation Protocol}
\label{appendix:annotation}

\subsection{LLM Classification Prompt (Localness Dimension Part)}

\begin{lstlisting}
You are given a question, a supporting context passage, and an answer --- all derived from a local news article. Your task is to identify **1 to 5 dimensions** from the list below that best describe the **main localness themes** expressed in the QA pair.

---
**Question:** {question}  
**Context:** {context}  
**Answer:** {answer}  
---

**INSTRUCTIONS**:
1. From the **dimension list** below, select **at least 1 and at most 5** dimensions that best characterize what this QA pair is about.
   - Only choose dimensions that are clearly relevant to the content.
   - If unsure, select the most general applicable dimension.
   - Do **not** default to the first few options: **read and consider the full list** before deciding.
2. You **must copy each dimension name exactly as written** below.
   - No paraphrasing, abbreviations, or extra characters.
   - Any deviations will result in invalid responses.
3. **Rank the selected dimensions by relevance**, placing the most relevant one first.
4. **Output format**:
   - One dimension per line.
   - No numbering, no bullet points, no explanation.
   - Output must contain **only valid dimension names**, exactly as listed.

---
**DIMENSION LIST** (unordered):
1. Cultural Understanding Dimension: Involves knowledge of symbolic, linguistic, and behavioral expressions that define local identity. This includes understanding local customs, language varieties, and cultural practices such as cuisine and social norms that shape how people interact and express belonging.  
2. Environmental Cognition Dimension: Captures mental representations of the physical and ecological characteristics of a place. This dimension focuses on how people understand the local geography, natural environment, and ecological systems through observation, learning, and reflection.  
3. Local Knowledge Dimension: Encompasses accumulated, context-specific information that informs decision-making and place literacy. This includes understanding historical developments, insider tips, changes over time, wayfinding skills, and the ability to offer localized recommendations.  
4. Place Interaction Dimension: Reflects embodied experience and direct physical engagement with the natural and built environment. This involves a lived connection with ecological features, comfort navigating local spaces, and sensory familiarity with landscapes and natural elements.  
5. Temporal Presence Dimension: Represents sustained residence or early-life connection that embeds individuals in the timeline of a place. This dimension emphasizes the depth of familiarity, emotional investment, and continuity afforded by being born in, growing up in, or living long-term in a location.  
6. Emotional Connection Dimension: Highlights affective bonds and the emotional significance of a place in personal life. It captures feelings of comfort, identity alignment, and deep attachment that make a place feel like ``home'' and contribute to a stable sense of belonging.  
7. Social and Community Engagement Dimension: Centers on interpersonal relationships, civic involvement, and contribution to communal life. This includes participating in events, engaging with local institutions, forming strong local networks, and expressing care through long-term commitments and civic action.  
\end{lstlisting}

\section{Appendix 9: Non-numerical QA Model Comparison}
\label{appen: Model_Comparation}

\begin{table}[H]
\centering
\begin{tabular}{lccccc}
\hline
\textbf{Model} & \textbf{GPT Judge Acc.} & \textbf{95\% CI} & \textbf{Effect Size (d)} & \textbf{Ans Rate} & \textbf{Conf Corr} \\
\hline
\textbf{Proprietary Models} & & & & &  \\
GPT-4.1 & 47.0 & [46.1, 47.7] & -- & 100.0 & 0.294 \\
GPT-4o & 32.8 & [32.6, 34.2] & \textbf{-0.412**} & 99.6 & 0.434 \\
Gemini-2.5-Pro & 52.5 & [51.9, 52.7] & \textbf{+0.485**} & 100.0 & -0.090 \\
Claude-Sonnet-4 & 39.7 & [39.3, 43.1] & \textbf{-0.203**} & 100.0 & 0.347 \\
\hline
\textbf{Web-Augmented} & & & & &  \\
GPT-4.1+Web & 35.6 & [35.2, 38.0] & \textbf{-0.289**} & 92.9 & 0.369 \\
Gemini-2.5-Pro+Grounding & \textbf{56.8} & [55.8, 58.6] & \textbf{+0.485**} & 91.7 & 0.211 \\
\hline
\textbf{Open-Source Models} & & & & &  \\
Qwen3-235B-A22B & 27.3 & [26.7, 31.3] & \textbf{-0.573**} & 99.3 & 0.228 \\
Qwen3-30B-A3B & 28.0 & [27.4, 31.0] & \textbf{-0.581**} & 99.7 & 0.247 \\
\hline
\end{tabular}
\caption{Statistical comparison for non-numerical QA tasks. Effect sizes are relative to GPT-4.1. \textbf{**}$p < 0.001$ under Bonferroni correction.}
\label{tab:statistical_analysis_non_num}
\end{table}

\section{Appendix 10: Numerical QA Model Comparison}

\begin{table}[H]
\centering
\begin{tabular}{lcccc}
\hline
\textbf{Model} & \textbf{Accuracy} & \textbf{95\% CI} & \textbf{Effect Size (d)} & \textbf{Ans Rate}\\
\hline
\textbf{Proprietary Models} & & & &  \\
GPT-4.1 & 6.2 & [4.9, 7.5] & -- & 100.0 \\
GPT-4o & 6.2 & [4.8, 7.6] & 0.000 & 39.8  \\
Gemini-2.5-Pro & 12.8 & [10.7, 14.9] & \textbf{+0.452**} & 100.0  \\
Claude-Sonnet-4 & 7.1 & [5.6, 8.6] & +0.078 & 97.3  \\
\hline
\textbf{Web-Augmented} & & & &   \\
GPT-4.1+Web & \textbf{15.5} & [13.1, 17.9] & \textbf{+0.623**} & 92.0  \\
Gemini-2.5-Pro+Grounding & 12.8 & [10.6, 15.0] & \textbf{+0.452**} & 100.0  \\
\hline
\textbf{Open-Source Models} & & & &   \\
Qwen3-235B-A22B & 6.6 & [5.0, 8.2] & +0.035 & 77.0  \\
Qwen3-30B-A3B & 2.2 & [1.2, 3.2] & \textbf{-0.342**} & 100.0  \\
\hline
\end{tabular}
\caption{Statistical comparison for numerical QA tasks. Effect sizes are relative to GPT-4.1. \textbf{**}$p < 0.001$ under Bonferroni correction.}
\label{tab:statistical_analysis_num}
\end{table}

\section{Appendix 11: Generation Success Rates}
\label{appen: generation_attempts}

Success rates calculated against respective base populations for each generation round, See Table~\ref{tab:complete_pipeline_stats}.

\begin{table}[H]
\centering
\small
\begin{tabular}{@{}lcccc@{}}
\hline
\textbf{Source/Attempt} & \textbf{Success Rate} & \textbf{QA Pairs} & \textbf{Base} & \textbf{Success (\%)} \\
\hline
\textbf{Census Data} & 100\% & 6,120 & 6,120 & 100.0 \\
\hline
Reddit + News Initial & -- & 8,240 & 9,107 & 87.6 \\
First Regeneration & 45.2\% & 527 & 9,107 & 93.2 \\
Second Regeneration & 22\% & 141 & 9,107 & 94.7 \\
Third Regeneration & 9.4\% & 47 & 9,107 & 95.2 \\
\hline
\textbf{Non-Census Subtotal} & -- & \textbf{8,662} & \textbf{9,107} & \textbf{95.1} \\
\textbf{All Sources Combined} & -- & \textbf{14,782} & \textbf{15,227} & \textbf{96.9} \\
\hline
\end{tabular}
\caption{Success rates calculated against respective base populations.}
\label{tab:complete_pipeline_stats}
\end{table}

\section{Appendix 12: Metric Details}
\label{appendix:metric_details}

\begin{table}[H]
\small
\centering
\begin{tabular}{lc}
\hline
\textbf{Metric} & \textbf{Purpose} \\
\hline
\textbf{Answer Rate} & Proportion of non-``I don't know'' responses, reflecting model engagement. \\
\textbf{Exact Match (EM)} & Measures strict string equality after normalization. \\
\textbf{ROUGE-1 F1} & Measures surface-level similarity tolerant of paraphrasing. \\
\textbf{Semantic Match} & Embedding-based cosine similarity using \texttt{text-embedding-3-small}. \\
\textbf{Numerical Accuracy Score} & Correct if its relative error is under 2\% of the gold value. \\
\textbf{GPT Judge Accuracy} & Binary correctness classification by GPT-4o-mini. \\
\textbf{GPT Judge Confidence} & Log-probability of GPT-4o-mini’s correctness token (``Correct''/``Incorrect''). \\
\textbf{Model Self-Confidence} & Scalar confidence reported by the model for its own answer. \\
\textbf{Confidence Correlation} & Pearson correlation between model confidence and GPT Judge accuracy. \\
\hline
\end{tabular}
\caption{Evaluation metrics used in \textsc{LocalBench}, including correctness, semantic similarity, numeric accuracy, and confidence-based measures.}
\label{tab:metrics}
\end{table}

\section{Appendix 13: Additional Implementation Details}
\label{appendix:implementation}

\paragraph{Reddit Data Processing:}
\begin{enumerate}
\item Extract posts and comments using PRAW with rate limiting (1 request/second)
\item Filter content by score threshold (posts: $\geq$ 5, comments: $\geq$ 3)
\item Remove deleted/removed content and moderator posts
\item Anonymize usernames and remove personal identifiers
\item Combine post text with top-50 comments for context
\end{enumerate}

\paragraph{News Data Processing:}
\begin{enumerate}
\item Filter NELA-Local articles by county-level geographic tags
\item Remove duplicate articles (cosine similarity $\geq$ 0.9)
\item Extract article text and publication metadata
\item Verify county association through location entity recognition
\item Sample articles to balance geographic distribution
\end{enumerate}

\paragraph{Census Data Processing:}
\begin{enumerate}
\item Download datasets from varies resources (See Table~\ref{tab:localness-metrics})
\item Standardize indicator names and units across data sources
\item Remove all counties with missing value
\item Validate data consistency across multiple census tables
\item Generate metadata descriptions for each indicator
\end{enumerate}




\section{Appendix 14: Census Metrics and Sources}
\label{appendix: metrics-localness-source}

\scriptsize
\begin{longtable}{p{2.0cm}p{2.5cm}p{6.5cm}p{5.0cm}}
\caption{Localness metric taxonomy with metric definitions and data sources.}
\label{tab:localness-metrics}\\
\toprule
\textbf{Domain} & \textbf{Dimension} & \textbf{Metrics} & \textbf{Data Source} \\
\midrule
\endfirsthead
\caption[]{Localness metric taxonomy with metric definitions and data sources.} \\
\toprule
\textbf{Domain} & \textbf{Dimension} & \textbf{Metrics} & \textbf{Data Source} \\
\midrule
\endhead
\midrule
\multicolumn{4}{r}{{Continued on next page}} \\
\midrule
\endfoot
\bottomrule
\endlastfoot
Cognitive & Cultural & In 2018, the number of nonemployer establishments in accommodation and food services per 1,000 residents in this county & U.S. Census Bureau Nonemployer Statistics Table NS1800NONEMP 2018 \\
\midrule
Cognitive & Cultural & In 2022, the number of residents in this county who spoke a language other than English at home & U.S. Census Bureau ACS Table DP02 \\
\midrule
Cognitive & Cultural & In 2022, the number of residents in this county who spoke English less than 'very well' at home & U.S. Census Bureau ACS Table DP02 \\
\midrule
Cognitive & Cultural & In 2020, the percentage of Southern Baptist Convention adherents among total adherents in this county & US Religion Census 2020 Group detail data by nation, state, county and metro \\
\midrule
Cognitive & Environmental & In 2018, the number of nonemployer establishments in mining, quarrying, and oil and gas extraction per 1,000 residents in this county & U.S. Census Bureau Nonemployer Statistics Table NS1800NONEMP 2018 \\
\midrule
Cognitive & Environmental & In 2022, the percentage of cropland fertilized in this county & USDA National Agricultural Statistics Service \\
\midrule
Cognitive & Knowledge & The change in multifamily building permits from 2021 to 2022 in this county & U.S. Census Bureau Building Permits Survey \\
\midrule
Cognitive & Knowledge & In 2018, the number of nonemployer establishments in information industries per 1,000 residents in this county & U.S. Census Bureau Nonemployer Statistics Table NS1800NONEMP 2018 \\
\midrule
Cognitive & Knowledge & As of 2024, the number of historic preservation properties with local significance in this county & National Register of Historic Places \\
\midrule
Cognitive & Knowledge & In 2018, the number of nonemployer establishments in professional, scientific, and technical services per 1,000 residents in this county & U.S. Census Bureau Nonemployer Statistics Table NS1800NONEMP 2018 \\
\midrule
Cognitive & Knowledge & In 2018, the number of nonemployer establishments in educational services per 1,000 residents in this county & U.S. Census Bureau Nonemployer Statistics Table NS1800NONEMP 2018 \\
\midrule
Cognitive & Knowledge & In 2018, the number of nonemployer establishments in administrative, support, and waste management and remediation services per 1,000 residents in this county & U.S. Census Bureau Nonemployer Statistics Table NS1800NONEMP 2018 \\
\midrule
Cognitive & Knowledge & In 2022, the number of public libraries in this county & Public Libraries Survey (PLS) by the Institute of Museum and Library Services (IMLS) \\
\midrule
Cognitive & Knowledge & In 2022, the mean travel time to work for residents of this county & U.S. Census Bureau ACS Table S0801 \\
\midrule
Cognitive & Knowledge & In 2022, the percentage of workers who use public transportation to work in this county & U.S. Census Bureau ACS Table S0801 \\
\midrule
Physical & Place-Interaction & In 2022, the percentage of employed residents living in this county who worked within their county of residence & U.S. Census Bureau ACS Table S0801 \\
\midrule
Physical & Place-Interaction & In 2018, the number of nonemployer establishments in agriculture, forestry, fishing, and hunting per 1,000 residents in this county & U.S. Census Bureau Nonemployer Statistics Table NS1800NONEMP 2018 \\
\midrule
Physical & Temporal & In 2022, the number of residents in this county who were born in the United States and in their state of residence & U.S. Census Bureau ACS Table DP02 \\
\midrule
Physical & Temporal & In 2022, the percentage of the population in this county identifying as Native American & U.S. Census Bureau ACS Table DP05 \\
\midrule
Physical & Temporal & In 2018, the number of nonemployer establishments in arts, entertainment, and recreation per 1,000 residents in this county & U.S. Census Bureau Nonemployer Statistics Table NS1800NONEMP 2018 \\
\midrule
Physical & Temporal & In 2022, the number of residents in this county who had lived in the same house or apartment for more than five years & U.S. Census Bureau ACS Table S0701 \\
\midrule
Physical & Temporal & In 2022, the median move‑in year of householders in owner‑occupied units in this county & U.S. Census Bureau ACS Table B25039 \\
\midrule
Physical & Temporal & In 2022, the percentage of occupied housing units that were owner‑occupied in this county & U.S. Census Bureau ACS Table DP04 \\
\midrule
Physical & Temporal & In 2022, the number of native residents in this county who moved to their current residence before 2010 & U.S. Census Bureau ACS Table DP02 \\
\midrule
Relational & Emotional & In 2022, the number of residents in this county who reported 'American' ancestry & U.S. Census Bureau ACS Table DP02 \\
\midrule
Relational & Emotional & In 2022, the percentage of residents in this county who were Hispanic or Latino (of any race) & U.S. Census Bureau ACS Table DP05 \\
\midrule
Relational & Emotional & In 2022, the ethnolinguistic fractionalization index of residents in this county & U.S. Census Bureau ACS Table DP05 \\
\midrule
Relational & Emotional & In 2022, the percentage of residents in this county with zero components of social vulnerability & Community Resilience Estimates Datasets \\
\midrule
Relational & Social/Community & In 2018, the number of nonemployer establishments in health care and social assistance per 1,000 residents in this county & U.S. Census Bureau Nonemployer Statistics Table NS1800NONEMP 2018 \\
\midrule
Relational & Social/Community & In the 2020 presidential election, the total number of votes cast in this county & County Presidential Election Returns 2000-2020 \\
\midrule
Relational & Social/Community & In 2022, the percentage of owner‑occupied housing units with a mortgage in this county & ACS Table DP05 \\
\midrule
Relational & Social/Community & In 2018, the density of nonemployer businesses per 1,000 residents in this county & U.S. Census Bureau Nonemployer Statistics Table NS1800NONEMP 2018 \\
\midrule
Relational & Social/Community & In 2022, the average size of married‑couple households in this county & U.S. Census Bureau ACS Table DP02 \\
\midrule
Relational & Social/Community & In 2022, the average household size in this county & U.S. Census Bureau ACS DP04 \\
\end{longtable}

\end{document}